%% file: main.tex
\pdfoutput=1
\documentclass[11pt]{article}

\usepackage[final]{acl}

\usepackage{times}
\usepackage{latexsym}

\usepackage[T1]{fontenc}
\usepackage[utf8]{inputenc}

\usepackage{microtype}

\usepackage{inconsolata}

\usepackage{amsmath, amssymb}
\usepackage{multirow, multicol}
\usepackage{booktabs, graphicx}
\usepackage{subfigure}
\usepackage{marvosym}
\usepackage[linesnumbered,ruled,vlined]{algorithm2e}

\DeclareMathOperator*{\argmax}{arg\,max}

\title{P-ICL: Point In-Context Learning for Named Entity Recognition \\ with Large Language Models}

\author{Guochao Jiang$^\dag$, Zepeng Ding$^\dag$, Yuchen Shi$^\dag$, Deqing Yang$^{\dag\ddag\textrm{\Letter}}$\\
$^\dag$School of Data Science, Fudan University, Shanghai, China\\
$^\ddag$Shanghai Key Laboratory of Data Science, Shanghai, China\\
$^\dag$\texttt{\{gcjiang22, zpding22, ycshi21\}@m.fudan.edu.cn} \\
$^\ddag$\texttt{yangdeqing@fudan.edu.cn}\\
}

\begin{document}
\maketitle
\begin{abstract}
In recent years, the rise of large language models (LLMs) has made it possible to directly achieve named entity recognition (NER) without any demonstration samples or only using a few samples through in-context learning (ICL). However, standard ICL only helps LLMs understand task instructions, format and input-label mapping, but neglects the particularity of the NER task itself. In this paper, we propose a new prompting framework \textbf{P-ICL} to better achieve NER with LLMs, in which some point entities are leveraged as the auxiliary information to recognize each entity type. With such significant information, the LLM can achieve entity classification more precisely. To obtain optimal point entities for prompting LLMs, we also proposed a point entity selection method based on K-Means clustering. Our extensive experiments on some representative NER benchmarks verify the effectiveness of our proposed strategies in P-ICL and point entity selection.\footnote{Our code is available at \url{https://github.com/jiangguochaoGG/P-ICL}.}
\end{abstract}

\input{Introduction.tex}
\input{Related_work}
\input{Preliminary}

\input{Method}
\input{Experiment}

\section{Conclusion}
In this paper, we introduce P-ICL, a framework that employs point entities to improve LLM in performing NER through ICL. By incorporating point entities, we equip LLMs with specific entity type information and entity classification, addressing the limitations of the standard ICL approach for NER. Our findings, based on six widely used NER datasets and experiments conducted with three renowned LLMs, confirm the efficacy of P-ICL.

\section*{Ethics Statement}
We hereby declare that all authors of this article are aware of and adhere to the provided ACL Code of Ethics and honor the code of conduct.
\paragraph*{Use of Human Annotations}
Human annotations are only used in methodological research at the beginning of the work, to assist in analyzing the feasibility of the proposed solution. Annotators consented to the use of data for research purposes. We ensure that the privacy of all annotators is protected throughout the annotation process, and all of them are adequately paid according to local standards. Human annotations are not applied during the evaluation of our method.
\paragraph*{Risks}
In this paper, all datasets are obtained from official sources. The datasets adopted have been anonymized and do not contain offensive information. However, we cannot guarantee that the datasets do not contain socially harmful or toxic language.

\section*{Limitations}
We focus solely on evaluating the P-ICL framework's performance with fixed point entities. For LLMs, having fixed point entities may not always be ideal for varying inputs. This limitation highlights the potential for future work on dynamically choosing point entities depending on the input sentence.

% \section*{Acknowledgements}
% This work was supported by the Chinese NSF Major Research Plan (No.92270121), Shanghai Science and Technology Innovation Action Plan (No.21511100401).

% Bibliography entries for the entire Anthology, followed by custom entries
%\bibliography{anthology,custom}
% Custom bibliography entries only
\bibliography{custom}

\newpage\appendix

\input{Appendix}

\end{document}

%% file: Introduction.tex
\section{Introduction}
\begin{figure}[!t]
    \centering
    \subfigure[Standard In-Context Learning]{\includegraphics[width=0.45\textwidth]{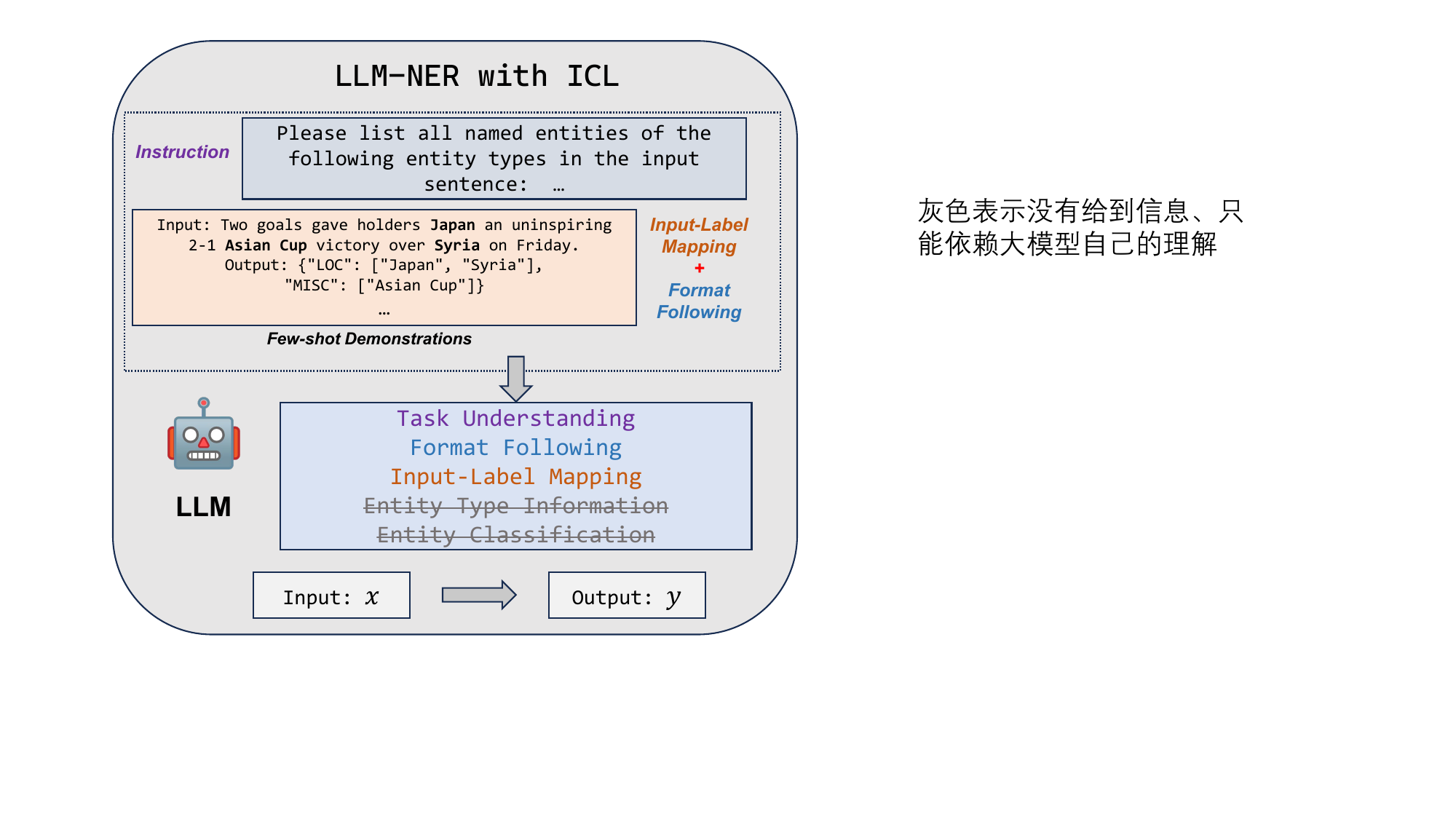}}
    \subfigure[Point In-Context Learning (P-ICL)]{\includegraphics[width=0.45\textwidth]{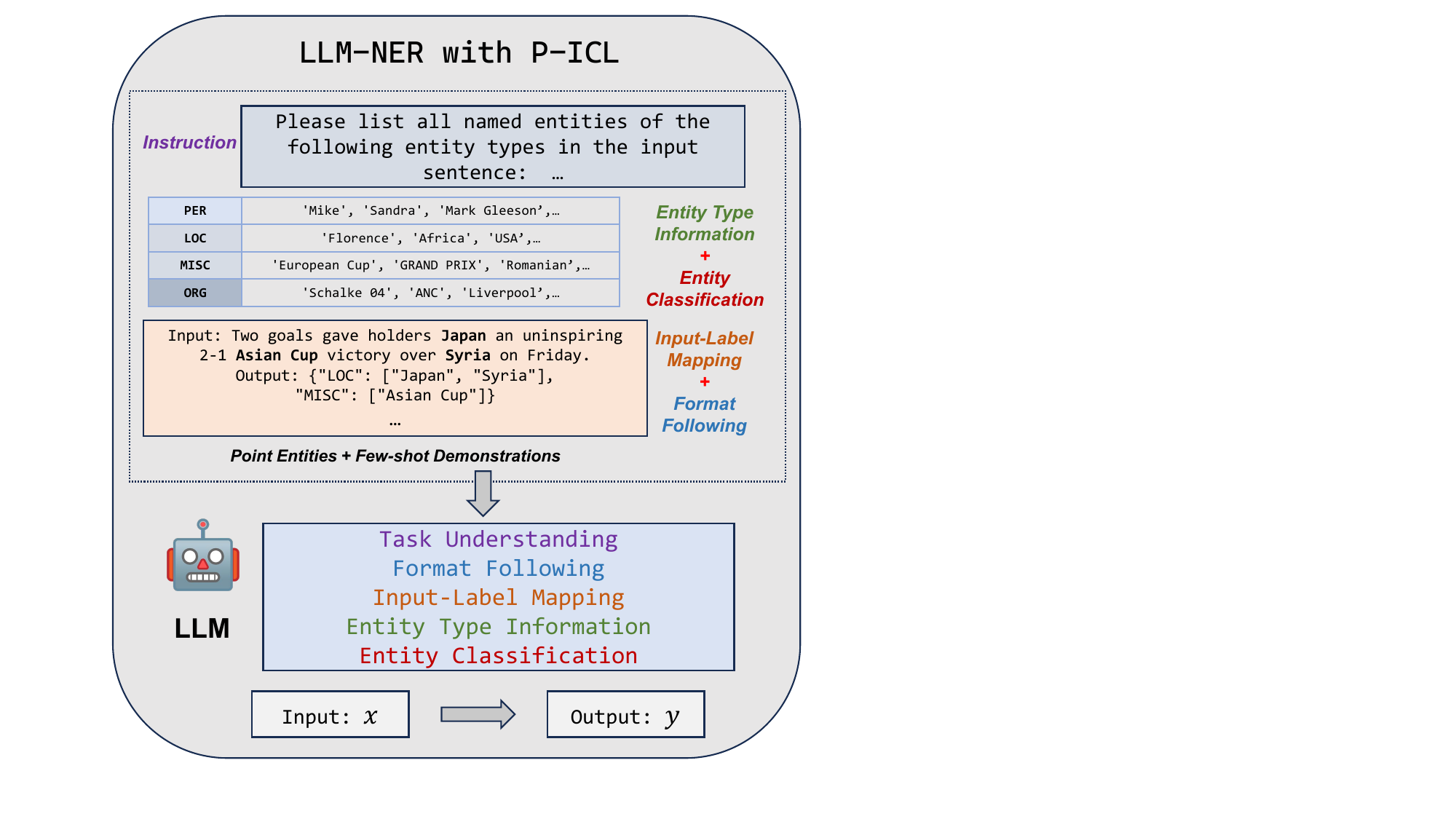}}
    \caption{The overview of LLM-based NER using the standard ICL and the P-ICL proposed in this paper. Standard ICL cannot provide LLM with sufficient entity type information and entity classification. Comparatively, P-ICL can make up for this shortcoming by providing some point entities per entity type.}
    \label{fig:icl_picl}
\end{figure}
% NER的背景
Named Entity Recognition (NER) \cite{NER} is a critical task of information extraction and plays a vital role in various downstream applications, including knowledge graph \cite{CNDB} construction, improving information retrieval systems \cite{banerjee2019information} and question-answering systems \cite{molla2006named}. The primary goal of NER is to precisely identify and classify the spans of entity mentions within a sentence into their corresponding entity types.

% LLM的NER，尤其是带ICL的few-shot NER
Large Language Models (LLMs), such as ChatGPT\footnote{\url{https://openai.com/blog/chatgpt}} and GPT-4 \cite{gpt-4}, have introduced new research directions in various natural language processing tasks. This shift is largely due to the in-context learning (ICL) abilities of LLMs, allowing them to perform tasks with only a few examples. With the rapid progress of LLMs, exploring their direct use or applying few-shot learning techniques to NER tasks is highly valuable. This strategy, which utilizes LLMs without extensive fine-tuning, offers a potential way to significantly lower the training costs associated with NER tasks.

% LLM使用ICL做NER的问题，ICL本身更多的是教会LLM下游任务的input-output范式来完成任务，而对于所做的NER任务本身的实体类型所代表的含义，则依赖LLM本身的知识
Existing research has often neglected the unique aspects of using LLMs for NER through ICL, especially when compared to their usage in other natural language processing (NLP) tasks. Previous study \cite{min-etal-2022-rethinking} has shown that improvements in LLMs' performance through ICL mainly come from input-label mapping and format following. Some recent works \cite{learning_icl_ner, good_examples, gpt-ner, instruct_uie} have explored the use of ICL for NER, but they ignored the role of entity examples explored in this paper. In fact, for NER, the information of entity type, along with entity classification, are the key factors to achieve overall task success. Unfortunately, such information has not yet been emphasized in previous ICL-based approaches.

To tackle this issue, we propose a new framework for NER in this paper, named \textbf{P-ICL} (\textbf{P}oint \textbf{I}n-\textbf{C}ontext \textbf{L}earning). Our framework equips the LLM with some representative entity instances for each entity type in the prompt, which provides significant information on entity type and entity classification. 
Compared with the demonstration examples in standard ICL's prompt, the representative entity instances added in P-ICL's prompt can be regarded as the data points in entity space, which are denoted as \textbf{Point} entities in this paper.
As shown in Figure \ref{fig:icl_picl}, providing these point entities for each entity type in the prompt greatly helps the LLM accurately understand the semantic distinctions between different entity types and their related domains. This enhancement is crucial for improving the NER performance of LLMs. Furthermore, designing effective point entity selection methods to select more representative point entities can help LLMs obtain entity type information and entity classification more effectively.

Our main contributions in this paper are summarized as follows.
\begin{enumerate}

\item We propose a novel framework P-ICL for LLMs to achieve NER. Our framework leverages point entities to provide the LLM with the significant information of entity type and entity classification, that is overlooked by standard ICL.
    
    \item The point entity selection method we proposed achieves better performance in the experiments compared with random selection, illustrating the significance of selecting more representative point entities for enhancing P-ICL's capability of NER.
    
    \item Our results of using three LLMs upon six representative NER benchmarks demonstrate the advantages of our proposed P-ICL over the standard ICL method.
\end{enumerate}

%% file: Related_work.tex
\section{Related Work}
% NER
% Large Language Model For NER or IE
% In-context learning Work

NER is an important task in the field of information extraction, which focuses on identifying named entities of interest from sentence-level or document-level text \cite{conll2003}. The existing NER solutions can be divided into three main categories: sequence labeling, span classification and generative methods. Using the sequence labeling approaches \cite{ratinov2009design, strakova2019neural, dai2020effective} split the text into tokens and the model assigns a label like BIO tagging to each token. Methods that use span classification \cite{wang2020pyramid, Yu2020NamedER} take the text into account in the form of the basic elements of a span, and then for the span decide whether it is an entity. Generative methods, on the other hand, rely on the overall framework of generative modeling to output the results of the final NER in the form of text generation \cite{cui2021template, yan2021unified}.

In recent years, LLMs have shown promising results on many NLP tasks including NER. \citet{wei2023zero} explored the direct use of ChatGPT for zero-shot named entity recognition, relation extraction and event extraction, and proposed a framework named ChatIE to interact with LLMs to accomplish the task of information extraction. Similarly, \citet{ji2023vicunaner} explored the performance of the open-source model Vicuna \cite{vicuna} on NER in the zero-shot and few-shot scenarios. \citet{hu2023zero} explored the NER capabilities of ChatGPT within a specific clinical domain, revealing the possibilities of applying generalized language models in specialized domains. \citet{xie-etal-2023-empirical} presented a systematic empirical investigation of zero-shot NER with LLM and adapted prevalent reasoning methods to NER. \citet{xie2023selfimproving} explored the possibility of boosting zero-shot NER with LLMs via self-improving. \citet{ToNER} introduced a generative NER framework named ToNER with entity type matching, which shows that entity mention is conducive to further improving NER performance. GoLLIE \cite{GoLLIE} employed an input-output format that includes schema definition, input text, and output annotations to perform IE tasks, but it ignored the role of point entity in this paper and the selection method for data examples.

In-context learning (ICL), as an emergent capability of LLMs, allows LLMs to quickly have the ability to adapt to downstream tasks given a few task examples in the prompt. The work that has been done focuses on the working mechanism of ICL and the factors that influence it \cite{min-etal-2022-rethinking, lyu-etal-2023-z, wei2023larger}. \citet{min-etal-2022-rethinking} found that the label space of the examples, the distribution of the input text and the overall form of the sequence were factors that influenced ICL performance. \citet{wang-etal-2023-label} delves into the mechanism of label words in ICL from the information flow perspective and introduces a re-weighting method to enhance the ICL performance of LLM based on this finding.

%% file: Preliminary.tex
\section{Preliminaries}
NER aims to extract entities from the given text and assign the correct entity type to the entities. Formally, we denote the input token sequence as $x = [x_1, x_2, \cdots, x_m]$. With LLM $f_\text{LLM}$, we denote the output token sequence as $y = f_\text{LLM}(x|\mathcal{I}) = [y_1, y_2, \cdots, y_n]$ under the instruction $\mathcal{I}$. Instruction $\mathcal{I}$ usually includes the entity types that the task focuses on such as PER, LOC, ORG and MISC in the CoNLL2003 dataset \cite{conll2003}.

LLMs can learn from demonstrations to improve the downstream performance with ICL ability. For NER, $i_t, t=1,\cdots,k$ represents the input text, $o_t, t=1,\cdots,k$ represents the extraction result, and $k$ represents the number of demonstrations. Given some demonstrations $(i_1, o_1), (i_2, o_2), \cdots, (i_k, o_k)$, LLMs can output results that are more consistent with the task instructions and these examples. Therefore, in the case of ICL, the extraction result of model $f_\text{LLM}$ for text $x$ is 
\begin{align}
y = f_\text{LLM}(x|\mathcal{I}, (i_1, o_1), (i_2, o_2), \cdots, (i_k, o_k))
\end{align}

To facilitate the analysis of the generated results, we stipulate that $o_t$ must be a parsable JSON result, where the key represents the entity type and the value is the entity list of the entity type corresponding to the key. %For example:

%\noindent\texttt{Input: He said further scientific study was required and if it was found that action was needed it should be taken by the European Union.}

%\noindent\texttt{Output: \{"ORG": ["European Union"]\}}\\

%\noindent\texttt{Input: Japan then laid siege to the Syrian penalty area and had a goal disallowed for offside in the 16th minute.}

%\noindent\texttt{Output: \{"LOC": ["Japan"], "MISC": ["Syrian"]\}}

%% file: Method.tex
\section{Methodology}
% LLM for NER with standard ICL
% point entity
% k-means method to get point entity
% LLM for NER with P-ICL

\begin{figure*}[!ht]
    \centering
    \includegraphics[width=0.9\textwidth]{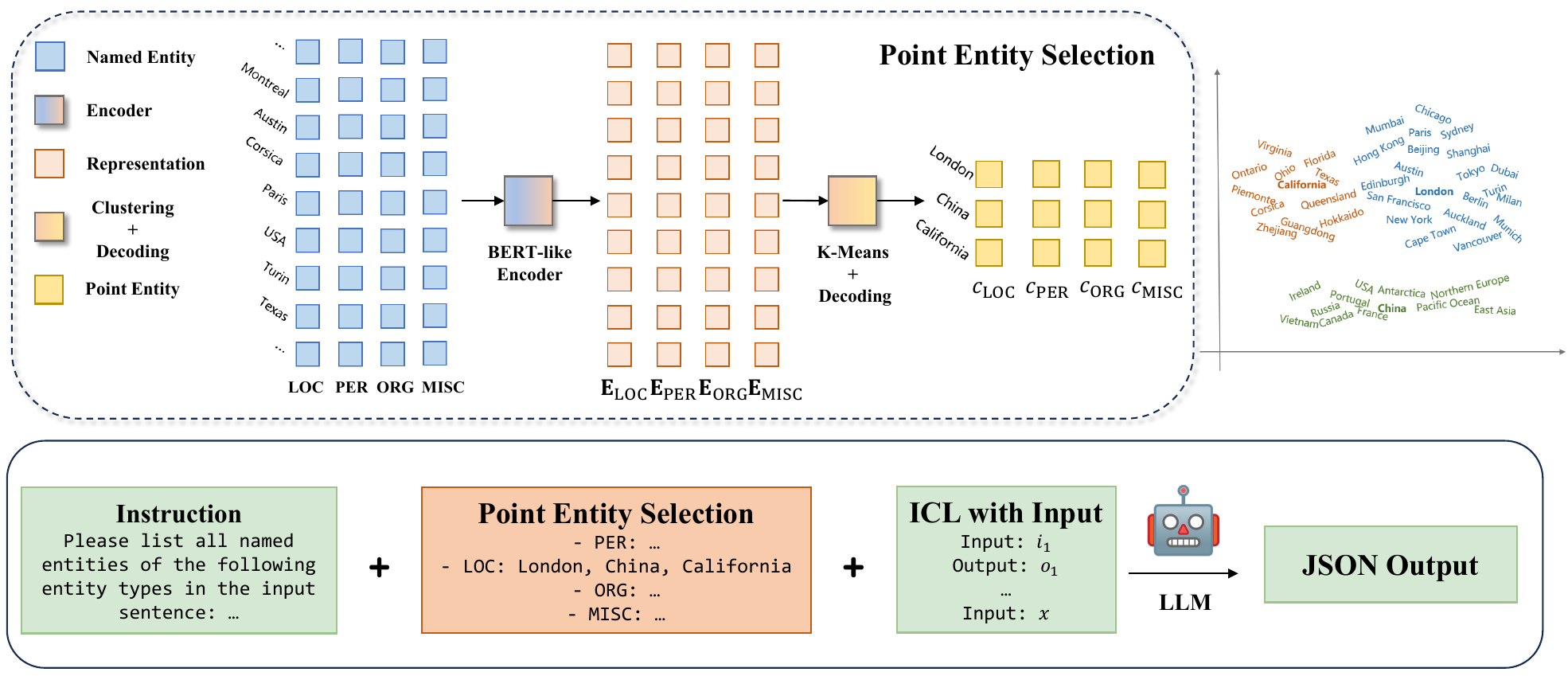}
    \caption{The overview of P-ICL with point entity selection. These entities are first given a vector representation using a BERT-like encoder. Then, the K-Means clustering method and nearest neighbor decoding strategy are used to identify the point entities for each entity type. The word cloud figure on the right displays the outcomes of this selection process. For instance, London, China, and California emerge as central entities in their clusters and are identified as point entities for LOC.}
    \label{fig:selection}
\end{figure*}

In this section, we initially discuss the standard ICL and point entity. Subsequently, we introduce our P-ICL and the method for selecting point entities. Figure \ref{fig:selection} illustrates the overview of our P-ICL pipeline, including the point entity selection process.

\subsection{Standard ICL}

In the previous discussion, we introduced a JSON schema aimed at facilitating the generation of structured and easily parsable outputs by LLM. Nonetheless, it is crucial to emphasize that without employing prompt engineering techniques, such as ICL, the outputs from LLM may not consistently adhere to the specified requirements.

ICL has been shown through empirical research to help LLMs learn the necessary input-output mapping relevant to specific downstream tasks. This strategy plays a key role in minimizing instances where LLM outputs fail to meet task requirements. In this study, which concentrates on the NER task, we employ the CoNLL2003 dataset as a case in point. We detail and apply particular prompts to carry out zero-shot and few-shot experiments under the ICL framework.

\vspace{0.5cm}

\noindent\textbf{Zero-shot}: \\
\small\noindent\texttt{Please list all named entities of the following entity types in the input sentence:} \\
\noindent\texttt{- PER} \\
\noindent\texttt{- ORG} \\
\noindent\texttt{- LOC} \\
\noindent\texttt{- MISC} \\ 
\noindent\texttt{You should output your results in the format \{"type": [entity]\} as a json.} \\
\noindent\texttt{Input: } $x$

\normalsize\noindent====================================

\noindent\textbf{Few-shot with ICL}: \\
\small\noindent\texttt{Please list all named entities of the following entity types in the input sentence:} \\
\noindent\texttt{- PER} \\
\noindent\texttt{- ORG} \\
\noindent\texttt{- LOC} \\
\noindent\texttt{- MISC} \\ 
\noindent\texttt{Here are some examples:} \\
\noindent\texttt{Input: } $i_1$ \\
\noindent\texttt{Output: } $o_1$ \\
\noindent\texttt{...} \\
\noindent\texttt{Input: } $i_k$ \\
\noindent\texttt{Output: } $o_k$ \\
\noindent\texttt{You should output your results in the format \{"type": [entity]\} as a json.} \\
\noindent\texttt{Input: } $x$

\normalsize\subsection{Point Entity}

In the approach described, the standard ICL method requires a certain number of examples to understand the details of downstream tasks. However, it does not fully capture the unique features of the NER task. As we know, two main factors affect NER performance: \textbf{contextual text} and \textbf{entity mention}. Typically, the contextual text provides extensive semantic information about a sentence, while the entity mention clearly identifies the potential entity. For LLMs, pre-training on a wide corpus gives the model a comprehensive understanding of natural language. As a result, many high-quality LLMs have strong text comprehension abilities. This is why LLMs perform well in various NLP tasks, even without specific training or in a few-shot ICL scenario.
% Thus, contextual text would not be a significant limitation for LLMs' performance in NER.
% This approach typically sees an improvement in model performance with an increase in the number of examples provided, following settings like 0-shot (no examples), 3-shot, 5-shot, 10-shot, and 20-shot. This trend suggests that as more examples are given, the model becomes better at understanding and performing the task at hand.

Entity mention, referring to how natural language identifies actual named entities, should be a key focus of NER for LLMs aiming for advanced text understanding. From a human intuition standpoint, entity mentions are usually quite specific. A major challenge in NER involves effectively defining the scope of each entity type in the downstream tasks. For LLMs, the strategies to address this challenge generally fall into two categories: providing detailed and clear definitions or presenting examples of entities. Embedding a specific number of entity instances for each entity type in the prompt can help LLMs capability discriminate the boundaries of relevant entity types, thereby boosting LLMs' performance on NER. Since this method only requires the presentations of entity mentions, which aligns well with the concept of a data point, we denote it as \textbf{Point} entity.

% For the methods that utilize neural networks, particularly those focusing on sequence labeling and span classification, model training relies heavily on the training data. This strategy aligns more closely with the previously mentioned latter solution. Unlike other approaches, these methods do not necessitate predefined definitions of entity types and they instead learn patterns directly from the training data.

% From the perspective of human cognition, the inclusion of entity mention examples can greatly enhance LLMs' capability of identifying the boundaries of various entity types. Often, these models cannot undergo fine-tuning in zero-shot and few-shot scenarios due to the prohibitive computational resource requirements, which exceed the capacities of most users. Standard ICL can empower general LLMs to acquire the knowledge about downstream tasks through the demonstration examples (about the task) in prompts. Similarly, embedding a specific number of entity instances for each entity type in the prompt can help LLMs capability discriminate the boundaries of relevant entity types, thereby boosting LLMs' performance on NER. Since this method only requires the presentations of entity mentions, which aligns well with the concept of a data point, we denote it as \textbf{Point} entity.

\subsection{P-ICL: Point ICL}

The concept of a point entity offers a novel approach to prompting LLM for NER task. By specifying the entities of interest through sufficient point entities, we can effectively communicate the types of entities to the LLM. This method, when compared to the standard ICL technique, allows for the inclusion of more training data for the LLM. Unlike standard ICL, which necessitates the full context, the point entity approach only requires the entity mention itself.

The ICL-like method that provides data information to LLM based on point entities is called \textbf{P-ICL}. Formally, assume that the entity type label set of the NER task is $\mathcal{T}$, and the number of entity types considered is $|\mathcal{T}|$. For the entity type $t \in \mathcal{T}$, there are $a_t$ point entities, respectively $p_{t, 1}, p_{t, 2}, \cdots, p_{t, a_t}$. In this paper, for the NER task, taking the CoNLL2003 dataset as an example, we consider the following prompts to conduct experiments under P-ICL:

\vspace{0.5cm}

\noindent\textbf{P-ICL}: \\
\small\noindent\texttt{Please list all named entities of the following entity types in the input sentence:} \\
\noindent\texttt{- PER: e.g. } $p_{\texttt{PER}, 1}$\texttt{,} $p_{\texttt{PER}, 2}$\texttt{,}$\cdots$\texttt{,}$p_{\texttt{PER}, a_\texttt{PER}}$ \\
\noindent\texttt{- ORG: e.g. } $p_{\texttt{ORG}, 1}$\texttt{,} $p_{\texttt{ORG}, 2}$\texttt{,}$\cdots$\texttt{,}$p_{\texttt{ORG}, a_\texttt{ORG}}$ \\
\noindent\texttt{- LOC: e.g. } $p_{\texttt{LOC}, 1}$\texttt{,} $p_{\texttt{LOC}, 2}$\texttt{,}$\cdots$\texttt{,}$p_{\texttt{LOC}, a_\texttt{LOC}}$ \\
\noindent\texttt{- MISC: e.g. } $p_{\texttt{MISC}, 1}$\texttt{,} $p_{\texttt{MISC}, 2}$\texttt{,}$\cdots$\texttt{,}$p_{\texttt{MISC}, a_\texttt{MISC}}$ \\
\noindent\texttt{Here are some examples:} \\
\noindent\texttt{Input: } $i_1$ \\
\noindent\texttt{Output: } $o_1$ \\
\noindent\texttt{...} \\
\noindent\texttt{Input: } $i_k$ \\
\noindent\texttt{Output: } $o_k$ \\
\noindent\texttt{You should output your results in the format \{"type": [entity]\} as a json.} \\
\noindent\texttt{Input: } $x$

\normalsize\subsection{Point Entity Selection} \label{sec:selection}

Building on the previous discussion, it's clear that the role of point entities is to provide LLMs with insights about different types of entities through mentions. Therefore, it makes sense to carefully select the right point entity. When point entities effectively represent the distribution of the entire dataset, LLMs can use their advanced natural language understanding to gain more information from inputs that are more representative, compared to when point entities are chosen randomly. This highlights the importance of careful selection of point entities to improve the performance of LLMs in processing and interpreting data, thus maximizing the information obtained from the inputs.

Recent advancements in Pre-trained Language Models (PLMs) like BERT \cite{bert} have provided many solutions for NLP tasks through representation learning. This approach mainly uses PLMs to extract relevant features from text for use in various tasks. In a given dataset, all the data in the training set can be used to help select representative entities. Since each entity can have a corresponding representation, choosing some representative entities aligns with unsupervised learning methods. In this scenario, the use of clustering algorithms \cite{cluster} proves to be an extremely effective technique.

By using PLMs alongside common clustering algorithms, it's possible to encode training entities related to a specific entity type to obtain their representations. Then, by clustering these representations, one can pinpoint central entities. The selection of a central entity, based on its position in the representation space, naturally reflects its representativeness. This method highlights the benefit of combining PLMs with clustering algorithms for a refined identification and selection of central entities. This enhances the model's capacity to generalize from representative examples in unsupervised settings.

Formally, let $\mathcal{E}_t = \left\{e_{t,1}, e_{t,2}, \cdots, e_{t,|\mathcal{E}_t|}\right\}$ be the entity set for entity type $t \in \mathcal{T}$. For each entity $e_t$ of entity type $t$, we can use the following method to obtain its corresponding representation:
\begin{align}
    \mathbf{e}_t = \text{Encoder}(e_t) \in \mathbb{R}^d,
\end{align}
where $d$ is the representation dimension, which generally depends on the encoder used. Let $\mathbf{E}_t = \left\{\mathbf{e}_{t,1}, \mathbf{e}_{t, 2}, \cdots, \mathbf{e}_{t, |\mathcal{E}_t|}\right\}$ be the corresponding entity representation set. Consistent with the above discussion, for entity type $t$, we use the following form to obtain $a_t$ entity centers with classic K-Means clustering algorithm:
\begin{align}
    (\mathbf{c}_{t,1}, \mathbf{c}_{t,2}, \cdots, \mathbf{c}_{t,a_t}) = \text{K-Means}(\mathbf{E}_t, a_t).
\end{align}

Then we need to convert the corresponding representation in the representation space into a string entity that can be used in prompt. For each entity center $\mathbf{c}_t$, we obtain the corresponding point entity $c_t$ by maximizing the similarity between the representation of the real entity and its entity center:
\begin{align}
    c_t = \argmax_{e \in \mathcal{E}_t} S\left(\text{Encoder}(e), \mathbf{c}_t\right),\label{eq:selection}
\end{align}
where the function $S$ measures the degree of similarity between two given elements, and is generally chosen as cosine similarity in most NLP tasks.

%% file: Experiment.tex
\section{Experiments}
% main result： CoNLL2003、WNUT2017、ACE2004、ACE2005数据集上的结果，包括模型GPT-3.5，GPT-4，Llama-2-70B，Mixtral-8x7B-Instruct，考虑设置ICL（0-shot，3-shot，5-shot，10-shot）、P-ICL（0-shot，3-shot，5-shot，10-shot）
In this section, we will evaluate the results of our proposed P-ICL method on some NER datasets, using different closed-source and open-source models. \footnote{Implementation details are in Appendix \ref{appendix:implementation}}

\subsection{Datasets}

\begin{table}[!t]
\centering
\resizebox{0.9\columnwidth}{!}{
\begin{tabular}{c|cccc}
\toprule
Dataset & Training & Training & Test & Test  \\
  Name  & Sen. \# & Ent. \# & Sen. \# & Ent. \#  \\ \hline
CoNLL2003  & 14,041 & 8,082 & 3,453  & 2,637 \\
WNUT2017 & 3,394 & 703 &  1,287 &  344 \\
ACE2004   & 6,202 & 10,090  & 812 & 1,692 \\ 
ACE2005   & 7,299 & 9,873  & 1,060 & 1,573 \\ 
JNLPBA   & 18,545 & 18,899  & 3,856 & 4,344 \\ 
BC5CDR   & 5,228 & 2,719  & 5,865 & 2,713 \\ 
\bottomrule
\end{tabular}
}
\caption{The datasets' statistics of sentence number and unique entity number in the train and test sets. In the zero-shot and few-shot scenarios, all the in-context examples and point entities are from the training set randomly.}
\label{table:datasets}
\end{table}

We conducted our experiments on the following NER benchmarks. These datasets' statistics are listed in Table \ref{table:datasets}.

\noindent\textbf{CoNLL2003} \quad CoNLL2003 \cite{conll2003} is a collection of news wire articles from the Reuters Corpus, which contains 4 entity types including \texttt{LOC}, \texttt{ORG}, \texttt{PER} and \texttt{MISC}.

\noindent\textbf{WNUT2017} \quad WNUT2017 \cite{wnut2017} focuses on unusual and previously-unseen entities in the context of emerging discussions, which contains 6 entity types including \texttt{person}, \texttt{location}, \texttt{corporation}, \texttt{product}, \texttt{creative-work} and \texttt{group}.

\noindent\textbf{ACE2004} and \textbf{ACE2005} \quad ACE2004 \cite{ace2004} and ACE2005 \cite{ace2005} are two nested named entity recognition datasets, which contain 7 entity types including \texttt{person}, \texttt{location}, \texttt{organization}, \texttt{geographical social political}, \texttt{weapon}, \texttt{facility} and \texttt{vehicle}. We followed the same dataset split setup as the previous work \cite{KatiyarC18, LinLHS19}.

\noindent\textbf{JNLPBA} \quad JNLPBA \cite{jnlpba} is a biomedical NER dataset, which contains 5 entity types including \texttt{DNA}, \texttt{protein}, \texttt{cell\_type}, \texttt{cell\_line} and \texttt{RNA}.

\noindent\textbf{BC5CDR} \quad BC5CDR \cite{bc5cdr} is a large annotated text corpus of human annotations of all chemicals, diseases and their interactions, which contains 2 entity types including \texttt{Chemical} and \texttt{Disease}.

\begin{table*}[ht]
\centering
\resizebox{1.0\textwidth}{!}{
\begin{tabular}{c|c|cccccc|c}
\hline
\multicolumn{2}{c|}{\textbf{Method}}              & \textbf{CoNLL2003} & \textbf{WNUT2017} & \textbf{ACE2004} & \textbf{ACE2005} & \textbf{JNLPBA} & \textbf{BC5CDR} & \textbf{Avg.} \\  \hline % GENIA BC5CDR
\multicolumn{2}{c|}{Vanilla} & 52.47 & 11.96 & 17.06 & 14.35 & 34.46 & 72.18 & 33.75 \\ \hline % -6
\multirow{4}{*}{Standard ICL} 
    & 3-shot & 65.15 & 13.49 & 32.27 & 32.85 & 45.06 & 71.54 & 43.39 \\ % -6
    & 5-shot & 67.24 & 14.21 & 36.70 & 37.10 & 46.77 & 72.07 & 45.68 \\ % -6
    & 10-shot & \underline{71.01} & \underline{14.68} & \underline{43.45} & \underline{43.50} & \underline{49.25} & \underline{72.24} & \underline{49.02} \\ % -6
    & 20-shot & \textbf{73.57} & \textbf{16.68} & \textbf{46.69} & \textbf{46.41} & \textbf{50.25} & \textbf{72.35} & \textbf{50.99} \\ \hline % -6
\multirow{8}{*}{P-ICL$_\text{random}$}    
    & 5+3-shot & 70.96 & 16.61 & 37.55 & 39.12 & 50.54 & 72.40 & 47.83 \\ % -4
    & 10+3-shot & 71.26 & 16.93 & 40.35 & 37.91 & 50.15 & 72.72 & 48.22 \\ % -4
    & 5+5-shot & 72.80 & 16.94 & 41.58 & 42.16 & 51.31 & 73.12 & 49.65 \\ % -4
    & 10+5-shot & 72.80 & 18.77 & 42.81 & 41.68 & 51.46 & 73.55 & 50.18 \\ % -4
    & 5+10-shot & 74.17 & 18.99 & 47.03 & 45.87 & 51.73 & 73.67 & 51.91 \\ % -4
    & 10+10-shot & 74.63 & 19.59 & 47.86 & 47.67 & 52.45 & 73.73 & 52.66 \\ % -4
    & 5+20-shot & \textbf{76.18} & \underline{20.74} & \underline{50.43} & \underline{49.28} & \underline{53.14} & \underline{74.10} & \underline{53.98} \\ % -4
    & 10+20-shot & \underline{76.07} & \textbf{21.06} & \textbf{50.50} & \textbf{50.33} & \textbf{53.61} & \textbf{74.62} & \textbf{54.37} \\ \hline % -4
\multirow{8}{*}{P-ICL$_\text{bert}$}
    & 5+3-shot & 75.48 & 22.10 & 43.73 & 42.56 & 53.61 & 77.40 & 52.48 \\
    & 10+3-shot & 76.43 & 21.28 & 44.91 & 42.56 & 53.92 & 76.52 & 52.60 \\
    & 5+5-shot & 76.36 & 22.02 & 46.47 & 46.24 & 55.07 & 78.33 & 54.08 \\
    & 10+5-shot & 76.70 & 23.19 & 47.70 & 46.44 & 54.96 & 77.35 & 54.39 \\
    & 5+10-shot & 78.95 & 21.51 & 52.11 & 52.39 & 56.70 & \underline{78.46} & 56.69 \\
    & 10+10-shot & 79.32 & 22.63 & 52.32 & 51.90 & 56.47 & 77.82 & 56.74 \\
    & 5+20-shot & \underline{80.67} & \underline{24.54} & \underline{54.45} & \textbf{54.13} & \textbf{56.91} & \textbf{78.49} & \underline{58.20} \\
    & 10+20-shot & \textbf{80.69} & \textbf{25.00} & \textbf{55.37} & \underline{53.61} & \underline{56.86} & 77.88 & \textbf{58.24} \\ \hline
\end{tabular}
}
\caption{Overall Experimental Results. Performance of LLaMA-3-70B across six datasets, with the best results highlighted in bold and the second-best underlined in each group.}
\label{table:llama-3-70b}
\end{table*}

\subsection{Overall Performance}

% For these datasets, we report the results of Mixtral 8x7B in the Table \ref{table:Mixtral-8x7B}, including the vanilla setting without ICL, 3-shot, 5-shot, 10-shot and 20-shot using ICL, and some P-ICL setting results. P-ICL$_\text{random}$ indicates that the point entities are randomly selected, and P-ICL$_\text{bert}$ indicates that the point entities are obtained by K-Means clustering based on the BERT-large model. The experimental results of models GPT-3.5 Turbo and LLaMA-2-70B are shown in the Appendix \ref{appendix:a}.

% From the tables, it is not difficult to find that on the four datasets, the performance of the model increases as the number of ICL demonstrations increases. However, the model performance does not necessarily improve when changing from 10 to 20 examples, which shows that the improvement brought by ICL to LLM is limited for NER. When the number of ICL demonstrations of LLM is given, the model performance of P-ICL is often higher than the model performance of ICL in the corresponding case. This result illustrates the advantages of our proposed P-ICL compared to the traditional use of only standard ICL. For situations where the number of ICL demonstrations is large, especially from 10-shot to 20-shot, the performance of the P-ICL model we proposed does not show a downward trend. For the same experimental setting, the model performance of P-ICL$_\text{bert}$ is higher than that of P-ICL$_\text{random}$. This finding reveals that our proposed point entity selection method is beneficial to further improving the performance of P-ICL.

We experimented on the six datasets with a variety of backbone LLMs, including LLaMA-3-70B, GPT-3.5 Turbo, and Mixtral 8x7B. The results for LLaMA-3-70B are documented in Table \ref{table:llama-3-70b}, while the outcomes for the other two models are detailed in Appendix \ref{appendix:a}. For each model, we employed different approaches, ranging from a zero-shot version without demonstrations, referred to as Vanilla, to few-shot versions, labeled ICL, which incorporate a range of example samples. For the P-ICL method proposed in this paper, we tested various selection techniques, including random selection and K-Means clustering-based selection, designated as P-ICL$_\text{random}$ and P-ICL$_\text{bert}$, respectively. For the ICL configurations, we evaluated the model's performance with different numbers of samples $b$, specifically $b=3, 5, 10, 20$. Moreover, in the P-ICL models, we experimented with the number of clusters $a$ per type, exploring $a=5, 10$. For instance, in the 5+3-shot experiment for P-ICL$_\text{bert}$, this denotes the selection of five entity examples per type and three NER samples for few-shot demonstrations. Based on the experimental results, we have the following observations:
\begin{enumerate}
    \item In the case of Standard ICL, we observed that the model's results improve with an increasing number of demonstrations until a certain threshold is reached. For example, there is no significant performance enhancement when scaling from 10-shot to 20-shot.
    \item When comparing Standard ICL with P-ICL$_\text{random}$ at equal numbers of demonstrations, the performance of P-ICL is consistently superior. This finding underscores the significance of introducing point entities for the NER task.
    \item Comparing P-ICL$_\text{random}$ with P-ICL$_\text{bert}$ under the same setting, the latter outperforms the former. This result validates the effectiveness of our proposed point entity selection method based on K-Means clustering in enhancing the performance of P-ICL.
\end{enumerate}
Furthermore, a comparison of experimental outcomes with $a=5$ and $a=10$ reveals minimal differences, hence, in the following section, we delve further into the analysis of the impact of point entity quantity.

\subsection{Effect of Point Entity Supervision}

\begin{table}[ht]
\centering
\resizebox{0.35\textwidth}{!}{
\begin{tabular}{c|cc}
\hline
ICL shot & \textbf{CoNLL2003} & \textbf{WNUT2017} \\ \hline
3 & 71.21 & 16.91 \\
5 & 72.95 & 17.54 \\
10 & 74.02 & 19.33 \\
20 & 76.03 & 21.57 \\ \hline
\end{tabular}
}
\caption{The performance of LLaMA-3-70B under point entity restriction on CoNLL2003 and WNUT2017 datasets, i.e., point entities can only come from ICL example labels.}
\label{table:supervision}
\end{table}

The source of point entities influences whether the model receives additional supervision beyond ICL. To investigate the impact of point entity supervision, we examine the performance of LLaMA-3-70B under point entity restriction on the CoNLL2003 and WNUT2017 datasets. Specifically, point entities can only originate from ICL example labels, as shown in Table \ref{table:supervision}. Compared to standard ICL results, the restricted P-ICL results consistently outperform standard ICL performance, demonstrating the advantage of P-ICL even without additional supervision. Additionally, when compared to P-ICL$_\text{random}$ with the same number of ICL shots, the restricted performance shows minimal difference. This further indicates that LLMs can efficiently learn task information from point entities, which will be further discussed in Section \ref{sec:number}.

\subsection{Effect of Point Entity Number}\label{sec:number}
% 对于固定icl数量的情况下，P-ICL表现随着picl数量的变化曲线
\begin{figure}
    \centering
    \includegraphics[width=0.43\textwidth]{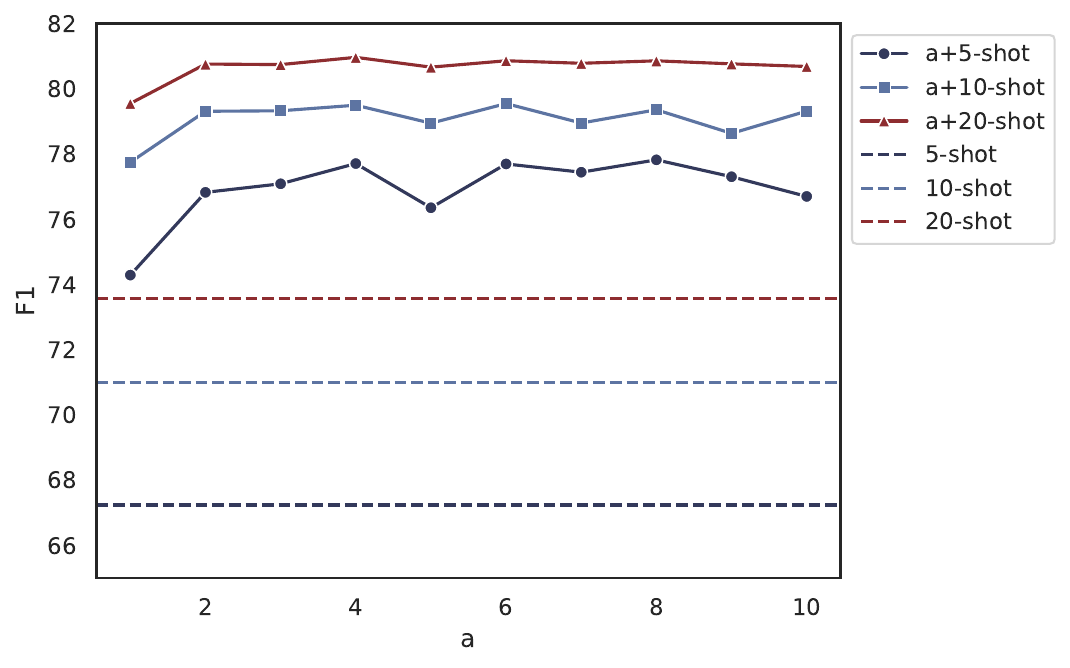}
    \caption{The impact of varying point entity numbers on model performance. Performance of LLaMA-3-70B on the CoNLL2003 dataset.}
    \label{fig:point_entity_number}
\end{figure}

% In order to further explore the impact of the number of point entities on the performance of P-ICL, we designed experiments on Mixtral 8x7B on the CoNLL2003 and WNUT2017 datasets. As shown in Figure \ref{fig:point_entity_number}, the two figures respectively show the change curves of the P-ICL performance of Mixtral 8x7B on the two datasets as the number of point entities increases. The line chart shows that the performance of P-ICL increases with the number of point entities when the number of point entities is small. Compared with standard ICL, the results of P-ICL under the same number of ICL demonstrations are higher than those of standard ICL. At the same time, the line chart results further reflect the problem that standard ICL has limited effects as the number of ICL demonstrations increases.

To delve deeper into how the quantity of point entities influences the performance of P-ICL, we conducted experiments on the CoNLL2003 dataset, with the results of LLaMA-3-70B displayed in Figure \ref{fig:point_entity_number}. For a fixed sample size $b=5, 10, 20$, we varied the number of point entities selected by BERT-large from each category, $a$, ranging from 1 to 10. The findings indicate that: (1) P-ICL outperforms standard ICL when comparing results under an equal number of ICL demonstrations. (2) With the same number of ICL demonstrations, P-ICL performance initially improves as the number of point entities increases. However, when the number of each type of point entity is high, the model's performance plateaus. This suggests that LLMs can efficiently learn task-relevant information from a given set of point entities, and a smaller number of point entities is sufficient.

\subsection{Effect of Point Entity Selection}
% 对于不同encoder进行的selection的影响，对于不同类选择point entity个数的影响

\begin{table*}[ht]
\centering
\begin{tabular}{c|cccc}
\hline
           & \textbf{BERT-large} & \textbf{RoBERTa-large} & \textbf{BGE-large} & \textbf{mxbai-embed-large} \\ \hline
\# Param. & 336M & 355M & 335M & 335M \\ \hline
5+3-shot   & 75.48 & 75.27 & \textbf{76.86} & \underline{75.55} \\
10+3-shot  & 76.43 & \underline{76.67} & \textbf{77.42} & 76.07 \\
5+5-shot   & 76.36 & 77.21 & \textbf{78.21} & \underline{77.46} \\
10+5-shot  & 76.70 & 77.80 & \textbf{78.86} & \underline{78.28} \\
5+10-shot  & 78.95 & 78.66 & \textbf{79.98} & \underline{79.30} \\
10+10-shot & 79.32 & 79.29 & \textbf{80.20} & \underline{79.65} \\
5+20-shot  & 80.67 & \underline{81.01} & \textbf{81.34} & 80.82 \\
10+20-shot & 80.69 & 80.44 & \underline{81.29} & \textbf{81.02} \\ \hline
\textbf{Avg.} & 78.06 & 78.29 & \textbf{79.27} & \underline{78.52} \\ \hline
\end{tabular}
\caption{Impact of different encoders on point entity selection. Results of LLaMA-3-70B on the CoNLL2003 dataset, with the best outcomes in bold and the second-best underlined.}
\label{table:encoder}
\end{table*}

% To explore the impact of Encoder in point entity selection, we selected two original BERT-like models (\textbf{BERT-large} \cite{bert} and \textbf{RoBERTa-large} \cite{roberta}) and two recent retrieval embedding models (\textbf{BGE-large} \cite{bge} and \textbf{mxbai-embed-large}\footnote{\url{https://www.mixedbread.ai/blog/mxbai-embed-large-v1}}) to conduct point entity selection experiments. As shown in the Table \ref{table:encoder} are the P-ICL performance results of different point entity selection models on the CoNLL2003 dataset with Mixtral 8x7B.

% Judging from the experimental results in Table \ref{table:encoder}, the point entity selection results using two BERT-like models are similar, and the point entity selection results using two point entity retrieval embedding models are similar. The individual results for each model are similar to the previous trends. At the same time, the point entity selection results using the retrieval embedding model are consistently better than the results of the two BERT-like models. This discovery inspired us to use an recent embedding model specifically for retrieval, which can provide better embeddings for entities, so that the point entities selected in point entity selection are more representative and provide more information gain for LLM.

To investigate the influence of point entity selection on performance outcomes, this section presents two sets of experiments. Firstly, we explore the impact of text encoders on results by utilizing different textual encoders for entity representation. Secondly, we examine the effects of varying the number of selected point entities for each entity center.

To examine the effects of different encoders, we selected four representative models: two classic BERT-like models (\textbf{BERT-large} \cite{bert} and \textbf{RoBERTa-large} \cite{roberta}), and two recent retrieval embedding models (\textbf{BGE-large} \cite{bge} and \textbf{mxbai-embed-large}\footnote{\url{https://www.mixedbread.ai/blog/mxbai-embed-large-v1}}). All of these models have a similar number of parameters. Employing these encoders for entity representation and performing point entity selection based on the methodology described in Section \ref{sec:selection}, we conducted experiments on the CoNLL2003 dataset. The results of LLaMA-3-70B, as depicted in Table \ref{table:encoder}, indicate that: (1) Utilizing BGE-large and mxbai-embed-large as encoder consistently outperforms BERT-like models, demonstrating that retrieval embedding models can yield more effective point entity selection results. (2) The outcomes for both BERT-like models are comparable, as are the results for the two retrieval embedding models. Overall, BGE-large exhibits superior performance to mxbai-embed-large, hence we have selected BGE-large as the encoder for our final model.

\begin{table}[ht]
\centering
\resizebox{0.35\textwidth}{!}{
\begin{tabular}{c|cc}
\hline
$\tau$ & P-ICL$_\tau$ & P-ICL$_\text{bert}$ + Extend \\ \hline
1 & 78.95 & 78.95 \\ \hline
2 & 79.56 & 79.40 \\
3 & 79.84 & 79.69 \\
4 & 79.57 & 79.85 \\
5 & 79.58 & 78.99 \\ \hline
\textbf{Avg.} & 79.50 & 79.38 \\ \hline
\end{tabular}
}
\caption{Impact of different $\tau$ values on performance of LLaMA-3-70B, with 5+10-shot setting on CoNLL2003. P-ICL$_\tau$ refers to the selection of $\tau$ most similar entities as point entities for each entity center, while P-ICL$_\text{bert}$ + Extend denotes the model extending P-ICL$_\text{bert}$ with randomly selected entities. }
\label{table:tau}
\end{table}

% In Eq. \ref{eq:selection}, we select the point entity by maximizing the similarity between the representation of the real point and entity center. This method is based on the idea that each entity center only needs one most representative entity. To test this idea, we designed the experiment as shown in the Table \ref{table:tau}. For each entity center, we select the $\tau$ entities that are most similar to it as the point entities corresponding to the entity center. When $\tau > 1$, one entity center will correspond to multiple point entities. In order to eliminate the influence of the number of point entities in the comparison experiments, we extend the corresponding number of point entities in the comparison experiment to keep the number of point entities consistent.

% Judging from the experimental results in Table \ref{table:tau}, as $\tau$ increases, the performance of P-ICL does not change significantly. At the same time, for each $\tau$, selecting multiple point entities does not have significantly higher results than one point entity. This shows that for each entity center, it is enough to select the closest point entity as a representative, and more point entities cannot bring more information gain to LLM.

In Equation \ref{eq:selection}, we select point entities by maximizing the similarity between the real entity's representation and the entity center. This approach is predicated on the notion that each entity center requires only one most representative entity. To validate this concept, we experimented with selecting a varying number of point entities for the same entity center, referred to as $\tau$. When $\tau > 1$, a single entity center corresponds to multiple point entities. Additionally, to eliminate the influence of the number of point entities in comparative experiments, we expanded the number of point entities by randomly selecting $(\tau - 1) \times a$ entities post-selection based on Equation \ref{eq:selection}. 
% , thus maintaining a consistent count of point entities across different experiments.

The experimental outcomes of LLaMA-3-70B for a 5+10-shot setting on the CoNLL2003 dataset are presented in Table \ref{table:tau}, where P-ICL$_\tau$ denotes the model selecting $\tau$ most similar entities as point entities for each entity center, and the corresponding P-ICL$_\text{bert}$ + Extend represents the model extending P-ICL$_\text{bert}$ with additional entities. The results reveal that: (1) Comparing P-ICL$_\tau$ for $\tau=1$ and $\tau>1$, selecting a larger $\tau$ does not lead to much better outcomes and may even result in a decline, suggesting that each entity center needs only the most representative entity, and selecting more similar entities could lead to redundancy due to excessive similarity. (2) There is no significant difference between the results of P-ICL$_\tau$ and P-ICL$_\text{bert}$ + Extend, implying that compared to point entities obtained through K-Means clustering, selecting more entities for each entity center or employing random extension does not yield better results.

%% file: Appendix.tex
\section{Implementation Details}\label{appendix:implementation}

We report the entity-level micro F1 scores in the following result tables and figures to compare the NER performance. To more comprehensively compare the P-ICL performance of existing LLMs, we selected three representative models at this stage, including GPT-3.5 Turbo, LLaMA-3-70B \cite{llama3modelcard} and Mixtral 8x7B \cite{mixtral}. The demonstrations used in ICL are selected randomly in the training set. For standard ICL and P-ICL experiments, due to a certain degree of randomness, the corresponding experimental results are the average of the F1 scores of 5 experiments. For experiments on open-source models, we use the vLLM \cite{vllm} framework for inference. In order to maintain the stability and consistency of model generation, we use the greedy generation to control the outputs of LLMs. We conducted our experiments on eight NVIDIA A800 80GB PCIe GPUs.

\section{Overall Performance}
\label{appendix:a}
The results for GPT-3.5 Turbo and Mixtral 8x7B are documented in Table \ref{table:gpt-3.5-turbo} and Table \ref{table:Mixtral-8x7B}.

\begin{table*}[ht]
\centering
\resizebox{1.0\textwidth}{!}{
\begin{tabular}{c|c|cccccc|c}
\hline
\multicolumn{2}{c|}{\textbf{Method}}              & \textbf{CoNLL2003} & \textbf{WNUT2017} & \textbf{ACE2004} & \textbf{ACE2005} & \textbf{JNLPBA} & \textbf{BC5CDR} & \textbf{Avg.}\\ \hline
\multicolumn{2}{c|}{Vanilla} & 62.49 & 8.70 & 17.99 & 14.35 & 38.83 & 68.30 & 35.11 \\ \hline % -4
\multirow{4}{*}{Standard ICL} 
    & 3-shot & 64.95 & 10.55 & 20.30 & 19.84 & 42.88 & 67.47 & 37.67 \\ % -4
    & 5-shot & 66.47 & 10.92 & 22.48 & 22.32 & 43.53 & 68.17 & 38.98\\ % -4
    & 10-shot & \textbf{67.65} & \textbf{11.51} & \textbf{23.94} & \underline{25.42} & \textbf{46.09} & \underline{68.67} & \textbf{40.55} \\ % -4
    & 20-shot & \underline{66.35} & \underline{10.39} & \underline{23.90} & \textbf{27.05} & \underline{45.05} & \textbf{69.12} & \underline{40.31} \\ \hline % -6
\multirow{8}{*}{P-ICL$_\text{random}$}    
    & 5+3-shot & 66.80 & 16.61 & 24.31 & 24.57 & 45.55 & 67.10 & 40.82 \\ % -2
    & 10+3-shot & 65.70 & 14.93 & 23.59 & 24.19 & 45.38 & 66.41 & 40.03 \\ % -2
    & 5+5-shot & 67.56 & 16.95 & 25.49 & 27.01 & 46.83 & 67.64 & 41.91 \\ % -2
    & 10+5-shot & 66.58 & 16.10 & 25.85 & 26.51 & 46.05 & 67.40 & 41.42 \\ % -2
    & 5+10-shot & 67.66 & 16.83 & 28.18 & 28.95 & 48.00 & 69.10 & 43.12 \\ % -2
    & 10+10-shot & 68.27 & 18.51 & 26.64 & 30.02 & 48.01 & 68.33 & 43.30 \\ % -2
    & 5+20-shot & \underline{69.31} & \textbf{19.36} & \textbf{29.10} & \textbf{32.05} & \underline{48.99} & \textbf{69.60} & \textbf{44.73} \\ % -2
    & 10+20-shot & \textbf{69.58} & \underline{18.40} & \underline{28.31} & \underline{31.99} & \textbf{49.16} & \underline{69.28} & \underline{44.45} \\ \hline % -2
\multirow{8}{*}{P-ICL$_\text{bert}$}    
    & 5+3-shot & 68.87 & 19.10 & 26.34 & 25.57 & 48.17 & 69.73 & 42.96 \\
    & 10+3-shot & 69.47 & 17.99 & 26.88 & 25.36 & 47.85 & 68.34 & 42.65 \\
    & 5+5-shot & 70.33 & 16.70 & 27.23 & 29.12 & 48.39 & 70.42 & 43.70 \\
    & 10+5-shot & 70.83 & 19.41 & 26.74 & 28.09 & 48.65 & 69.18 & 43.82 \\
    & 5+10-shot & 71.04 & 18.18 & 29.66 & 31.03 & 50.22 & \textbf{71.27} & 45.23 \\
    & 10+10-shot & 71.12 & \underline{20.98} & 29.11 & 30.45 & \underline{51.04} & 70.36 & 45.51 \\
    & 5+20-shot & \textbf{72.58} & \textbf{21.09} & \textbf{31.17} & \textbf{32.69} & \textbf{51.06} & \underline{71.26} & \textbf{46.64} \\
    & 10+20-shot & \underline{71.80} & 20.23 & \underline{30.53} & \underline{32.19} & 50.86 & 71.10 & \underline{46.12} \\ \hline
\end{tabular}
}
\caption{Overall Experimental Results. Performance of GPT-3.5 Turbo across six datasets, with the best results highlighted in bold and the second-best underlined in each group.}
\label{table:gpt-3.5-turbo}
\end{table*}

\begin{table*}[!ht]
\centering
\resizebox{1.0\textwidth}{!}{
\begin{tabular}{c|c|cccccc|c}
\hline
\multicolumn{2}{c|}{\textbf{Method}}              & \textbf{CoNLL2003} & \textbf{WNUT2017} & \textbf{ACE2004} & \textbf{ACE2005} & \textbf{JNLPBA} & \textbf{BC5CDR} & \textbf{Avg.} \\ \hline
\multicolumn{2}{c|}{Vanilla} & 37.38 & 6.31 & 19.01 & 17.26 & 21.85 & 54.63 & 26.07 \\ \hline % -4
\multirow{4}{*}{Standard ICL} 
    & 3-shot & 45.93 & 8.76 & 28.22 & 25.42 & 26.95 & \underline{57.69} & 32.16 \\ % -4
    & 5-shot & 48.10 & 10.11 & 30.54 & 27.90 & 28.16 & 57.68 & 33.75 \\ % -4
    & 10-shot & \underline{51.00} & \textbf{12.56} & \underline{32.42} & \textbf{30.94} & \underline{29.38} & \textbf{57.83} & \textbf{35.69} \\ % -4
    & 20-shot & \textbf{51.12} & \underline{11.37} & \textbf{32.93} & \underline{29.78} & \textbf{32.08} & 56.38 & \underline{35.61} \\ \hline % ? -4
\multirow{8}{*}{P-ICL$_\text{random}$}    
    & 5+3-shot & 46.37 & 10.62 & 31.09 & 28.79 & 25.89 & \underline{59.11} & 33.65 \\ % -2
    & 10+3-shot & 45.85 & 11.00 & 30.74 & 28.91 & 25.46 & 56.91 & 33.15 \\ % -2
    & 5+5-shot & 48.87 & 11.67 & 31.13 & 31.33 & 28.15 & 58.84 & 35.00 \\ % -2
    & 10+5-shot & 48.53 & 12.64 & 33.26 & 31.06 & 27.88 & 58.30 & 35.28 \\ % -2
    & 5+10-shot & 53.09 & 13.49 & 35.92 & 34.57 & 29.86 & \textbf{59.45} & 37.73 \\ % -2
    & 10+10-shot & 53.07 & 13.09 & 35.37 & 35.04 & \underline{31.33} & 57.96 & 37.64 \\ % -2
    & 5+20-shot & \underline{56.50} & \textbf{14.85} & \textbf{37.02} & \textbf{36.69} & \textbf{32.08} & 58.28 & \textbf{39.24} \\ % -2
    & 10+20-shot & \textbf{56.67} & \underline{14.26} & \underline{36.01} & \underline{35.22} & 30.44 & 57.99 & \underline{38.43} \\ \hline % -2
\multirow{8}{*}{P-ICL$_\text{bert}$}    
    & 5+3-shot & 50.59 & 12.54 & 32.18 & 31.27 & 27.87 & \textbf{62.14} & 36.10 \\
    & 10+3-shot & 48.59 & 13.00 & 33.67 & 30.69 & 29.38 & 58.17 & 35.58 \\
    & 5+5-shot & 51.81 & 14.55 & 36.08 & 34.03 & 29.21 & \underline{61.36} & 37.84 \\
    & 10+5-shot & 50.02 & 13.75 & 37.12 & 33.59 & 31.89 & 59.23 & 37.60 \\
    & 5+10-shot & 56.18 & 15.36 & 36.77 & 36.78 & 32.40 & 61.10 & 39.77 \\
    & 10+10-shot & 54.42 & 15.06 & 37.82 & 36.14 & \underline{32.99} & 59.79 & 39.37 \\
    & 5+20-shot & \textbf{59.12} & \textbf{16.92} & \underline{39.38} & \underline{38.60} & \textbf{35.05} & 60.41 & \textbf{41.58} \\
    & 10+20-shot & \underline{58.70} & \underline{16.19} & \textbf{41.18} & \textbf{38.34} & 32.26 & 59.18 & \underline{40.98} \\ \hline
\end{tabular}
}
\caption{Overall Experimental Results. Performance of Mixtral 8x7B across six datasets, with the best results highlighted in bold and the second-best underlined in each group.}
\label{table:Mixtral-8x7B}
\end{table*}